\documentclass[runningheads]{llncs}
\usepackage[final,year=2024,ID=549]{configs/eccv}

\usepackage{configs/eccvabbrv}
\usepackage{graphicx}
\usepackage{booktabs}
\usepackage{comment}

\usepackage[accsupp]{axessibility}  

\usepackage[pagebackref,breaklinks,colorlinks]{hyperref}

\usepackage{orcidlink} 
\usepackage{graphics}
\usepackage{overpic}
\usepackage{overpic} %
\usepackage{color}
\usepackage{multirow}
\usepackage{paralist}
\usepackage{makecell}
\usepackage{soul} %
\usepackage{pifont}%
\newcommand{\xmark}{\ding{55}}%
\usepackage{tabularx}
\usepackage{svg}
\usepackage{amsmath}
\usepackage{subcaption}
\usepackage{graphicx}
\usepackage[export]{adjustbox}
\usepackage{tikz}
\usetikzlibrary{matrix,calc}
\usepackage[normalem]{ulem}
\usepackage{pifont}
\usepackage[dvipsnames]{xcolor}

\usepackage{appendix}

\newcommand{\modelname}{\textcolor{black}{TeSMo}\xspace}

\newcommand{\projectURL}{\href{https://research.nvidia.com/labs/toronto-ai/tesmo}{\tt\textit{https://research.nvidia.com/labs/toronto-ai/tesmo}}}

\definecolor{citecolor}{HTML}{0071bc}

\newcommand{\tref}[1]{Tab.~\ref{#1}}

\newcommand{\fref}[1]{Fig.~\ref{#1}}

\definecolor{jtcolor}{RGB}{0,0,255}

\definecolor{todocolor}{RGB}{255,0,00}

\newcommand\todo[1] {\emph{\textcolor{todocolor}{TODO: #1}}}

\definecolor{turquoise}{cmyk}{0.65,0,0.1,0.3}
\definecolor{purple}{rgb}{0.65,0,0.65}
\definecolor{dark_green}{rgb}{0, 0.5, 0}
\definecolor{orange}{rgb}{0.8, 0.6, 0.2}
\definecolor{red}{rgb}{0.8, 0.2, 0.2}
\definecolor{darkred}{rgb}{0.6, 0.1, 0.05}
\definecolor{blueish}{rgb}{0.0, 0.3, .6}
\definecolor{light_gray}{rgb}{0.7, 0.7, .7}
\definecolor{pink}{rgb}{1, 0, 1}
\definecolor{greyblue}{rgb}{0.25, 0.25, 1}
\definecolor{forestgreen}{rgb}{0.0, 0.2, 0.13}
\definecolor{darkolivegreen}{rgb}{0.33, 0.42, 0.18}

\newcommand{\qheading}[1]{\noindent\textbf{#1}.}
\newcommand{\mheading}[1]{\medskip\noindent\textbf{#1}.}

\begin{document}

\pagestyle{headings}
\mainmatter

\title{Generating Human Interaction Motions in Scenes with Text Control}
%


\author{Hongwei Yi\inst{1,2} \and
Justus Thies\inst{2,3} \and
Michael J. Black\inst{2} \and \\ 
Xue Bin Peng\inst{1,4} \and
Davis Rempe\inst{1}}

\institute{
NVIDIA
\and
Max Planck Institute for Intelligent Systems, Tubingen, Germany 
\and
Technical University of Darmstadt
\and 
Simon Fraser University
}

\authorrunning{H.Yi et al.}


\maketitle




\vspace{-0.35in}
\begin{figure}
    \centerline{
     \includegraphics[width=1\textwidth]{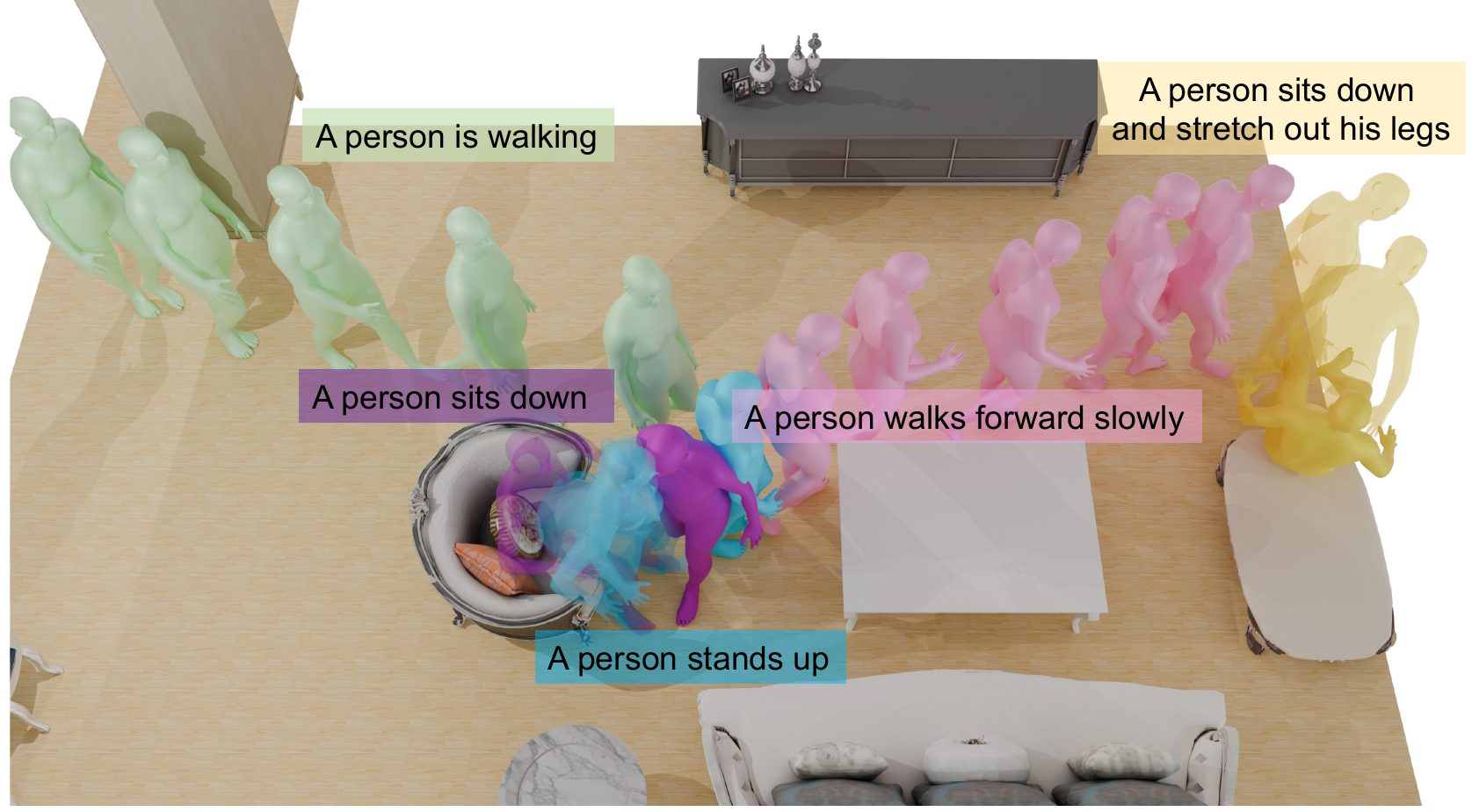}}
     \vspace{-2mm}
    \caption{We present \modelname, a method for generating diverse and plausible human-scene interactions from text input. Given a 3D scene, \modelname generates scene-aware motions, such as walking in free space and sitting on a chair. Our model can be easily controlled using textual descriptions, start positions, and goal positions.}
    \label{fig:teaser}
\end{figure}
\vspace{-0.5in}

\begin{abstract}
We present \modelname, a method for text-controlled scene-aware motion generation based on denoising diffusion models. Previous text-to-motion methods focus on characters in isolation without considering scenes due to the limited availability of datasets that include motion, text descriptions, and interactive scenes. Our approach begins with pre-training a scene-agnostic text-to-motion diffusion model, emphasizing goal-reaching constraints on large-scale motion-capture datasets. We then enhance this model with a scene-aware component, fine-tuned using data augmented with detailed scene information, including ground plane and object shapes. To facilitate training, we embed annotated navigation and interaction motions within scenes. The proposed method produces realistic and diverse human-object interactions, such as navigation and sitting, in different scenes with various object shapes, orientations, initial body positions, and poses. Extensive experiments demonstrate that our approach surpasses prior techniques in terms of the plausibility of human-scene interactions, as well as the realism and variety of the generated motions. Code will be released upon publication of this work at \projectURL.

\end{abstract}

\section{Introduction}
Generating realistic human movements that can interact with 3D scenes is crucial for many applications, ranging from gaming to embodied AI. 
For example, character animators for games and films need to author motions that successfully navigate through cluttered scenes and realistically interact with target objects, while still maintaining artistic control over the style of the movement.
One natural way to control style is through text, e.g., ``skip happily to the chair and sit down''. 
Recently, diffusion models have shown remarkable capabilities in generating human motion from user inputs. Text prompts~\cite{22iclr_mdm,22arxiv_motiondiffuse} let users control style, while methods incorporating spatial constraints enable more fine-grained control, such as specifying desired joint positions and trajectories~\cite{xie2023omnicontrol, priormdm, karunratanakul2023gmd}.
However, these works have predominantly focused on characters in isolation, without considering environmental context or object interactions.

In this work, we aim to incorporate scene-awareness into user-controllable human motion generation models.
However, learning to generate motions involving scene interactions is challenging, even without text prompts. Unlike large-scale motion capture datasets that depict humans in isolation~\cite{AMASS:ICCV:2019}, datasets with paired examples of 3D human motion and scene/object geometry are limited.
Prior work uses small paired datasets without text annotations to train VAEs~\cite{21iccv_samp, 22eccv_couch, 19tog_nsm} or diffusion models~\cite{23arxiv_scenediffuser,Pi_2023_ICCV} that generate human scene interactions with limited scope and diversity.
Reinforcement learning methods are able to learn interaction motions from limited supervision \cite{23sig_physicalcsi,zhao2023dimos,lee2023locomotion}, and can generate behaviors that are not present in the training motion dataset. 
However, designing reward functions that lead to natural movements for a diverse range of interactions is difficult and tedious. 

To address these challenges, we introduce a method for Text-conditioned Scene-aware Motion generation, called \modelname. 
As shown in \cref{fig:teaser}, our method generates realistic motions that navigate around obstacles and interact with objects, while being conditioned on a text prompt to enable stylistic diversity.
Our key idea is to combine the power of general, but scene-agnostic, text-to-motion diffusion models with paired human-scene data that captures realistic interactions.
First, we pre-train a text-conditioned diffusion model~\cite{22iclr_mdm} on a diverse motion dataset with no objects (e.g., HumanML3D~\cite{22cvpr_humanml3d}), allowing it to learn a realistic motion prior and the correlation with text. 
We then fine-tune the model with an augmented scene-aware component that takes scene information as input, thereby refining motion outputs to be consistent with the environment.

Given a target object with which to interact and a text 
prompt describing the desired motion, we decompose the problem of generating a suitable motion in a scene into two components, \textit{navigation} (e.g., approaching a chair while avoiding obstacles) and \textit{interaction} (e.g., sitting on the chair). 
Both stages leverage diffusion models that are pre-trained on scene-agnostic data, then fine-tuned with an added scene-aware branch.
The \textit{navigation} model generates a pelvis trajectory that reaches a goal pose near the interaction object. 
During fine-tuning, the scene-aware branch takes, as input, a top-down 2D floor map of the scene and is trained on our new dataset containing locomotion sequences~\cite{AMASS:ICCV:2019} in 3D indoor rooms~\cite{fu20213d}.
The generated pelvis trajectory is then lifted to a full-body motion using motion in-painting~\cite{priormdm}.
Next, the \textit{interaction} model generates a full-body motion conditioned on a goal pelvis pose and a detailed 3D representation of the target object. 
To further improve generalization to novel objects, the model is fine-tuned using augmented data that re-targets interactions~\cite{21iccv_samp} to a variety of object shapes while maintaining realistic human-object contacts. 

Experiments demonstrate that our navigation approach outperforms prior work in terms of goal reaching and obstacle avoidance, while producing full-body motions on par with scene-agnostic diffusion models~\cite{xie2023omnicontrol,karunratanakul2023gmd}. 
Meanwhile, our interaction model generates motions with fewer object penetrations than the state-of-the-art approach~\cite{zhao2023dimos}, being preferred 71.9\% of the time in a perceptual study.
The central contribution of this work includes: (\textbf{1}) a novel approach to enable scene-aware and text-conditioned motion generation by fine-tuning an augmented model on top of a pre-trained text-to-motion diffusion model, (\textbf{2}) a method, \modelname, that leverages this approach for navigation and interaction components to generate high-quality motions in a scene from text, (\textbf{3}) data augmentation strategies for placing navigation and interaction motions with text annotations realistically in scenes to enable scene-aware fine-tuning.

\vspace{-2mm}
\section{Related Work}

\subsection{Scene-aware Motion Generation}
Motion synthesis in computer graphics has a rich history, encompassing areas such as locomotion~\cite{22cvpr_gamma, Agrawal_vandePanne_2016,Lee_Choi_Lee_2006, Kovar_Gleicher_Pighin}, human-scene/object interaction~\cite{Lee_Chai_Reitsma_Hodgins_Pollard_2002, taheri2021goal}, and dynamic object interaction~\cite{Corona_Pumarola_Alenyà_Moreno-Noguer_2019, li2023controllable, li2023object}.
We refer readers to an extensive survey \cite{zhu2023human} for an overview and focus on scene-aware motion generation in this section. 

A particular challenge in modeling scene-aware motion is the lack of paired, high-quality human-scene datasets.
One line of work \cite{wang2021scene, wang2021synthesizing} employs a two-stage method that first predicts the root path, followed by the full-body motion based on the scene and predicted path. However, these methods suffer from low-quality motion generation, attributed to the noise in the training datasets captured from monocular RGB-D videos \cite{19iccv_prox}.
Neural State Machine (NSM) \cite{19tog_nsm} proposes the use of phase labeling \cite{Holden_Komura_Saito_2017} and local expert networks \cite{Eigen_Ranzato_Sutskever_2013,Jacobs_Jordan_Nowlan_Hinton_1991,Yuksel_Wilson_Gader_2012} 
to generate high-quality object interactions, such as sitting and carrying, after training on a small human-object mocap dataset. Nonetheless, it struggles with recognizing walkable regions in 3D scenes, often failing to avoid obstacles. 
Therefore, later work in this vein requires using the A* algorithm for collision-free path planning~\cite{21iccv_samp}. These and related approaches~\cite{22eccv_couch,zhang2024roam} are moreover limited by the diversity of the small human-scene interaction datasets with no text annotations.

Various approaches ameliorate the data issue by creating synthetic data with captured~\cite{yi2022mime,ye2023summon} or generated~\cite{kulkarni2023nifty} motions placed in scenes heuristically. 
HUMANISE~\cite{wang2022humanise} does this for text-conditioned scene interactions, but rely entirely on short synthetic sequences for training, where the realism is limited by the data generation heuristics used.
The reinforcement learning (RL) approach DIMOS~\cite{zhao2023dimos} learns autoregressive policies to reach goal poses in a scene without requiring paired human-scene data for training, but still relies on A* and is constrained by the accuracy of goal pose generation \cite{COINS:ECCV:2022}. 
RL with physical simulators \cite{chao2021learning, 2022-TOG-ASE, 23sig_physicalcsi,xiao2023unified} has been used to produce physically plausible movements but faces challenges in generalizing across varied scenes and objects.
%

Unlike most prior works, our approach is text-conditioned and leverages a mix of both scene-agnostic and paired human-scene data. Pre-training is done with a diverse scene-agnostic dataset, while scene-aware fine-tuning uses motion data with scene context. For training, we adopt both synthetic data creation with real motions and data augmentation of real-world human-object interactions~\cite{21iccv_samp}.
\vspace{-3mm}
\subsection{Diffusion-Based Motion Generation}
Recently, diffusion models have demonstrated the ability to generate high-quality human motions, especially when conditioned on a text prompt~\cite{22iclr_mdm, 22arxiv_motiondiffuse,petrovich24stmc}. 
In addition to text, several diffusion models add spatial controllability. Some works~\cite{22iclr_mdm, priormdm} adopt image inpainting techniques to incorporate dense trajectories of spatial joint constraints into generated motions. OmniControl~\cite{xie2023omnicontrol} and GMD~\cite{karunratanakul2023gmd} allow control with sparse signals and a pre-defined root path, respectively.

A few diffusion works handle interactions with objects or scenes. 
TRACE~\cite{rempeluo2023tracepace} generates 2D trajectories for pedestrians based on a rasterized street map.
SceneDiffuser~\cite{23arxiv_scenediffuser} conditions generation on a full scanned scene point cloud, but motion quality is limited due to noisy training data~\cite{19iccv_prox}. 
Another approach~\cite{Pi_2023_ICCV} tackles single-object interactions through hierarchical generation of milestone poses followed by dense motion, but it lacks text control. 
A concurrent line of work enables text conditioning for single-object interactions~\cite{diller2023cghoi,peng2023hoi}, but they focus on humans manipulating dynamic objects rather than interactions in full scenes.


We leverage a pre-trained text-to-motion diffusion model~\cite{22iclr_mdm} and a fine-tuned scene-aware branch to enable both text controllability and scene-awareness with diffusion. 
We break motion generation into navigation and interaction with static objects by conditioning on 2D floor maps and 3D geometry, respectively, and create specialized human-scene data to enable diversity and quality.
\vspace{-2mm}
\section{Text-Conditioned Scene-Aware Motion Generation}
\subsection{Overview}
\begin{figure*}[ht!]
    \centerline{
    \includegraphics[width=\textwidth]{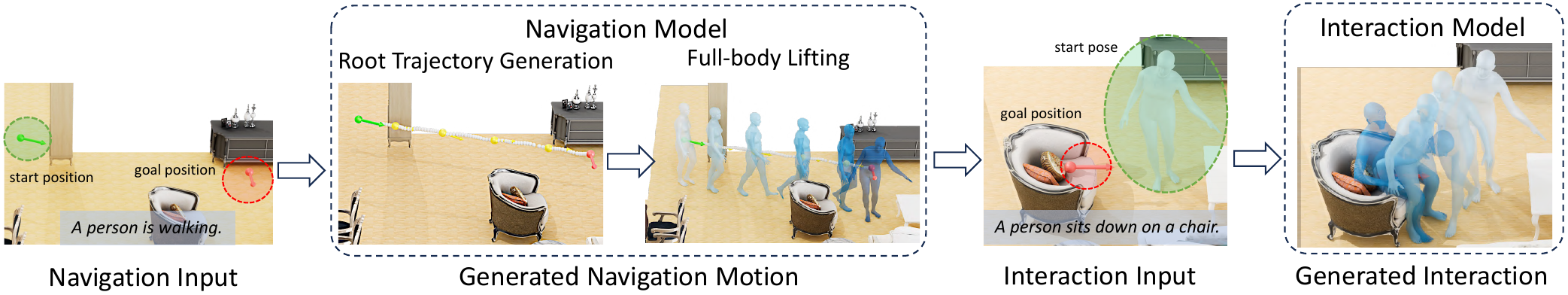}}
    \vspace{-2mm}
    \caption{Pipeline overview: given the start position (green arrow), goal position (red arrow), 3D scene, and text description, the navigation root trajectory is first generated and then the full-body motion is completed through in-painting. Subsequently, the interaction is generated from a start pose (i.e., the end pose from navigation), goal position, and the target object, enabling the generation of object-specific motion.
    }
    \vspace{-4mm}
    \label{fig:overview}
\end{figure*}

Given a 3D scene and a target interaction object, our goal is to generate a plausible human-scene interaction, where the motion style can be controlled by a user-specified text prompt.
Our approach decomposes this task into two components, \textit{navigation} and \textit{interaction}, as illustrated in \fref{fig:overview}. 
Both components are diffusion models that leverage a fine-tuning routine to enable scene-awareness without losing user controllability, as introduced in \cref{sec:background}.
To interact with an object, the character must first navigate to a location in the scene near the object, which is easily calculated heuristically or specified by the user, if desired. 
As described in \cref{t2loco}, we design a hierarchical \textit{navigation} model, which generates a root trajectory starting from an initial location that moves to the goal location while navigating around obstacles in the scene. The generated root trajectory is then lifted into a full-body motion using in-painting techniques~\cite{priormdm,xie2023omnicontrol}.
%
Since the navigation model gets close to the object in the first stage, for generating the actual object \emph{interaction}, we can focus on scenarios where the character is already near the object. This allows a one-stage motion generation model that directly predicts the full-body motion from the starting pose (\ie, the last pose of navigation), 
a goal pelvis pose, and the object (as detailed in Section~\ref{t2interact}).
\vspace{-2mm}
\subsection{Background: Controllable Human Motion Diffusion Models}
\label{sec:background}
\paragraph{Motion diffusion models.}
Diffusion models have been successfully used to generate both top-down trajectories~\cite{rempeluo2023tracepace} and full-body motions~\cite{22iclr_mdm,22arxiv_motiondiffuse}. 
These models generate motions by iteratively denoising a temporal sequence of $N$ poses (e.g., root positions or full-body joint positions/angles) $\mathbf{x} = \left[ \mathbf{x}^1, \dots, \mathbf{x}^N \right]$. 
During training, the model learns to reverse a forward diffusion process, which starts from a clean motion $\mathbf{x}_0 \sim q\left(\mathbf{x}_0\right)$, sampled from the training data, and after $T$ diffusion steps is approximately Gaussian $\mathbf{x}_{T} \sim \mathcal{N}(\mathbf{0}, \mathbf{I})$. 
Then at each step $t$ of motion denoising, the reverse process is defined as:
\begin{equation}
    p_\phi (\mathbf{x}_{t-1} | \mathbf{x}_t, \mathbf{c}) = \mathcal{N} \left ( \mathbf{x}_{t-1}; \boldsymbol{\mu}_\phi (\mathbf{x}_t, \mathbf{c}, t), \beta_t \mathbf{I} \right )
\end{equation}
where $\mathbf{c}$ is some conditioning signal (e.g., a text prompt), and $\beta_t$ depends on a pre-defined variance schedule. 
The denoising model $\boldsymbol{\mu}_\phi$ with parameters $\phi$ predicts the denoised motion $\hat{\mathbf{x}}_0$ from a noisy input motion $\mathbf{x}_t$~\cite{20nips_ddpm}. 
The model is trained by sampling a motion $\mathbf{x}_0$ from the dataset, adding random noise, and supervising the denoiser with a reconstruction loss $\left\|\mathbf{x}_0-\hat{\mathbf{x}}_0\right\|^2$.
\vspace{-2mm}
\paragraph{Augmented controllability.}
In the image domain, general pre-trained diffusion models are specialized for new tasks using an augmented ControlNet~\cite{zhang2023controlnet} branch, which takes in a new conditioning signal and is fine-tuned on top of the frozen base diffusion model. 
OmniControl~\cite{xie2023omnicontrol} adopts this idea to the human motion domain. For motion diffusion models with a transformer encoder architecture, they propose an augmented transformer branch that takes in kinematic joint constraints (e.g., pelvis or other joint positions) and at each layer connects back to the base model through a linear layer that is initialized to all zeros.

As described in \cref{t2loco,t2interact}, our key insight is to use an augmented control branch to enable scene awareness. We first train a strong scene-agnostic motion diffusion model to generate realistic motion from a text prompt, and then fine-tune an augmented branch that takes scene information as input (e.g., a 2D floor map or 3D geometry). This new branch adapts generated motion to be scene-compliant, while still maintaining realism and text controllability.
\vspace{-2mm}
\paragraph{Test-time guidance.}
At test time, diffusion models can be controlled to meet specific objectives through guidance.
We directly apply guidance to the clean motion prediction from the model $\hat{\mathbf{x}}_0$~\cite{rempeluo2023tracepace, ho2022video}. At each denoising step, the predicted $\hat{\mathbf{x}}_0$ is perturbed with the gradient of an analytic objective function $\mathcal{J}$ as $\tilde{\mathbf{x}}_0=\hat{\mathbf{x}}_0-\alpha \nabla_{\mathbf{x}_t} \mathcal{J}\left(\hat{\mathbf{x}}_0\right)$
where $\alpha$ controls the strength of the guidance and $\mathbf{x}_t$ is the noisy input motion at step $t$. 
The predicted mean $\boldsymbol{\mu}_\phi$ is then calculated with the updated motion prediction $\tilde{\mathbf{x}}_0$ as in \cite{rempeluo2023tracepace, ho2022video}. 
As detailed later, we define guidance objectives for avoiding collisions and reaching goals.
%

\vspace{-2mm}
\subsection{Navigation Motion Generation} \label{t2loco}

The goal of the navigation stage is for the character to reach a goal location near the target object using realistic locomotion behaviors that can be controlled by the user via text. 
We design a hierarchical method that first generates a dense root trajectory with a diffusion model, then leverages a powerful in-painting model~\cite{priormdm} to generate a full-body motion for the predicted trajectory. 
This approach facilitates accurate goal-reaching with the root-only model, while allowing diverse text control through the in-painting model.

\begin{figure*}[t]
    \centerline{
    \includegraphics[width=\textwidth]{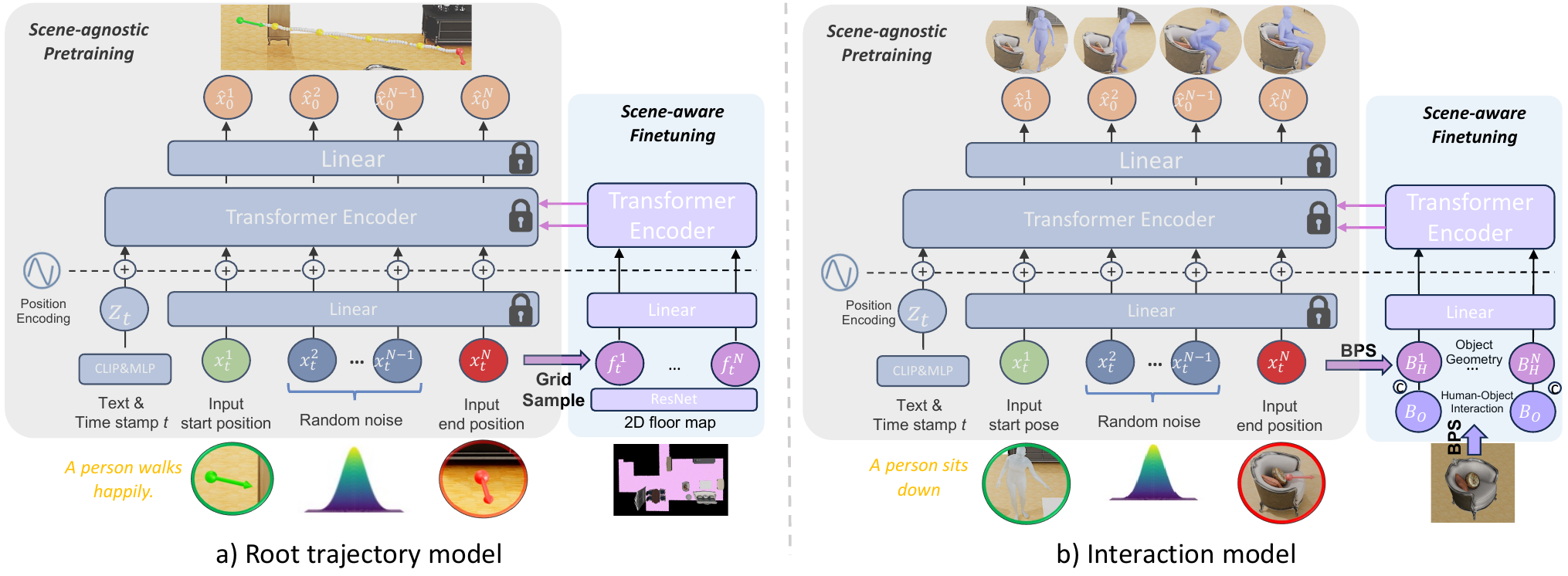}}
    \vspace{-2mm}
    \caption{
    Network architecture of the (\textbf{a}) root trajectory model and (\textbf{b}) interaction motion model. Initially, the base transformer encoder is trained on scene-agnostic motion data using start pose, target pose, and text as input. 
    Subsequently, a scene-aware component is fine-tuned, which incorporates the 2D floor map (a) or 3D object (b).
}
    \vspace{-5mm}
    \label{fig:arctecture}
\end{figure*}

\vspace{-2mm}
\paragraph{Root trajectory generation.}
Our root trajectory diffusion model, shown in \cref{fig:arctecture}(a), operates on motions where each pose is specified by $\mathbf{x}^n = \left[ x, y, z, \cos \theta, \sin\theta \right]_n$, with ($x,y,z$) being the pelvis position and $\theta$ the pelvis rotation, both of which are represented in the coordinate frame of the \emph{first} pose in the sequence.  
The model is conditioned on a text prompt along with starting and ending goal positions and orientations.
In contrast to the representation from prior work~\cite{22cvpr_humanml3d}, which uses relative pelvis velocity and rotation, our representation using absolute coordinates facilitates constraining the outputs of the model with goal poses. 

%
%
Inspired by motion in-painting models\cite{22iclr_mdm, priormdm}, given a start pose $\mathbf{s}$ and end goal pose $\mathbf{g}$, at each denoising step, we mask out the input $\mathbf{x}_t$ such that $\mathbf{x}^1_t = \mathbf{s}$ and $\mathbf{x}^N_t = \mathbf{g}$, thereby providing clean goal poses directly to the model.
To achieve this, a binary mask $\mathbf{m} = \left[ \mathbf{m}^1, \dots, \mathbf{m}^N \right]$ with the same dimensionality as $\mathbf{x}_t$ is defined, where $\mathbf{m}^1$ and $\mathbf{m}^N$ are a vector of 1's and all other $\mathbf{m}^n$ are 0's. 
During training, overwriting occurs with $\tilde{\mathbf{x}}_t = \mathbf{m} * \mathbf{x}_0 + (\mathbf{1} - \mathbf{m}) * \mathbf{x}_t$ where $*$ indicates element-wise multiplication and $\mathbf{x}_0$ is a ground truth root trajectory.
We then concatenate the mask with the overwritten motion $\left[ 
\tilde{\mathbf{x}}_t; \mathbf{m} \right]$ and use this as input to the model to indicate which frames have been overwritten.
%

At test time, goal-reaching is improved using a guidance objective $\mathcal{J}_{g} = ( \hat{\mathbf{x}}^N_0 - \mathbf{g} )^2$
 that measures the 
error between the end pelvis position and orientation of the predicted clean trajectory $\hat{\mathbf{x}}^N_0$ and the final goal pose. 

\vspace{-2mm}
\paragraph{Incorporating scene representation.}
The model as described so far is trained on a locomotion subset of the HumanML3D dataset \cite{22cvpr_humanml3d} to enable generating realistic, text-conditioned root trajectories. 
However, it will be entirely unaware of the given 3D scene. 
To take the scene into account and avoid degenerating the text-following and goal-reaching performance, we augment the base diffusion model with a control branch that takes a representation of the scene as input. 
This scene-aware branch is a separate transformer encoder that is fine-tuned on top of the frozen base model.
As input, we extract the walkable regions from the 3D geometry of the scene and project them to a bird's-eye view, yielding a 2D floor map $\mathcal{M}$. 
Following \cite{rempeluo2023tracepace}, a Resnet-18 \cite{he2015deep} encodes the map $\mathcal{M}$ as feature grid, and at denoising step $t$, each 2D projected pelvis position $(x,z) \in \boldsymbol{x}^n_t$ is queried in the feature grid $\mathcal{M}$ to get the corresponding feature $\mathbf{f}^n_t$. The resulting sequence of features $\mathbf{f}_t {=} \left[ \mathbf{f}^1_t, \dots, \mathbf{f}^N_t \right]$, along with the text prompt and noisy motion $\mathbf{x}_t$, become the input to the separated transformer branch.  
%

At test time, a collision guidance objective further encourages scene compliance.
This is defined as $\mathcal{J}_c = \text{SDF}(\hat{\mathbf{x}}_0, \mathcal{M})$ where $\text{SDF}$ calculates the 2D transform distance map from the 2D floor map, then queries the 2D distance value at each time step of the root trajectory. Positive distances, indicating pelvis positions outside the walkable region, are averaged to get the final loss.

\vspace{-2mm}
\paragraph{Scene-aware training and data.}
To train the scene-aware branch, it is important to have a dataset featuring realistic motions navigating through scenes with corresponding text prompts. 
For this purpose, we create the \textbf{Loco-3D-FRONT} dataset by integrating locomotion sequences from HumanML3D into diverse 3D environments from 3D-FRONT~\cite{fu20213d}. Each motion is placed within a different scene with randomized initial translation and orientation, following the methodology outlined in \cite{yi2022mime}, as depicted in \fref{fig:dataset}(a). 
Additionally, we apply left-right mirroring to both the motion and its interactive 3D scenes to augment the dataset~\cite{22cvpr_humanml3d}. This results in a dataset of approximately 9,500 walking motions, each motion accompanied by textual descriptions and 10 plausible 3D scenes on average, resulting in 95k locomotion-scene training pairs.

\vspace{-2mm}
\paragraph{Added control with trajectory blending.}
Our root trajectory diffusion model generates scene-aware motions and, unlike many prior works~\cite{21iccv_samp,zhao2023dimos}, does not require a navigation mesh to compute A*~\cite{hart1968formal} paths to follow. 
However, a user may want a character to take the shortest path to an object by following the A* path, or to control the general shape of the path by drawing a 2D route themselves.
To enable this, we propose to fuse an input 2D trajectory $\mathbf{p} \in \mathbb{R}^{N\times 2}$ with our model's predicted clean trajectory at every denoising step. 
%
%
At step $t$, we extract the 2D ($x,z$) components $\hat{\mathbf{p}}_0$ from the predicted root trajectory $\hat{\mathbf{x}}_0$ and interpolate them with the input trajectory $\tilde{\mathbf{p}}_0 = s*\hat{\mathbf{p}}_0 + (1-s)*\mathbf{p}$
where $s$ is the blending scale that controls how closely the generated trajectory matches the input. 
We then overwrite the 2D components of $\hat{\mathbf{x}}_0$ with $\tilde{\mathbf{p}}_0$ and continue denoising. 
This blending procedure ensures outputs roughly follow the desired path but still maintain realism inherent to the trained diffusion model.

\vspace{-2mm}
\paragraph{Lifting to full-body poses.}
To lift the generated pelvis trajectory to a full-body motion, we leverage the existing text-to-motion in-painting method PriorMDM~\cite{priormdm}, which takes a dense 2D root trajectory as input.
By using this strong model that is pre-trained for text-to-motion, we can effectively generate natural and scene-aware full-body motion, while offering diverse stylistic control through text.
%

\vspace{-3mm}
\subsection{Object-Driven Interaction Motion Generation} \label{t2interact} 
After navigation, the character has reached a location near the target object and next should execute a desired interaction motion. 
Due to the fine-grained relationship between the body and object geometry during interactions, we propose a single diffusion model to directly generate full-body motion, unlike the two-stage navigation approach from \cref{t2loco}.
%
\vspace{-2mm}
\paragraph{Interaction motion generation.}
The interaction motion model operates on a sequence of full-body poses and is shown in \cref{fig:arctecture}(b). 
Our pose representation extends that of HumanML3D~\cite{22cvpr_humanml3d} to add the absolute pelvis position and heading $(x,y,z,\cos\theta,\sin\theta)$, similar to our navigation model. Each pose in the motion is $\mathbf{x}^n = \left [  x,\; y,\; z,\; \sin{\theta},\; \cos{\theta},\; \dot{r}^a,\; \dot{r}^x,\; \dot{r}^z,\; r^y,\; \mathbf{j}^p,\; \mathbf{j}^v,\; \mathbf{j}^r,\; \mathbf{c}^f \right ]_n \in \mathbb{R}^{268}$ with $\dot{r}^a$ root angular velocity, $(\dot{r}^x, \dot{r}^z)$ root linear velocity, $r^y$ root height, $\mathbf{c}^f$ foot contacts, and $\mathbf{j}^p$, $\mathbf{j}^v$, $\mathbf{j}^r$ the local joint positions, velocities, and rotations, respectively. 

The model is conditioned on a text prompt 
along with a starting full-body pose (\ie, the final pose of the navigation stage) and a final goal pelvis position and orientation. 
The goal pelvis pose can usually be computed heuristically, but may also be provided by the user or predicted by another network~\cite{21iccv_samp}. 
The same masking procedure described in \cref{t2loco} is used to pass the start and end goals as input to the model.
At test time, we also use the same goal-reaching guidance to improve the accuracy of hitting the final pelvis pose.

\vspace{-2mm}
\paragraph{Object representation.}
The base interaction diffusion model is first trained on a dataset of interaction motions from HumanML3D and SAMP~\cite{21iccv_samp} without any objects, which helps develop a strong prior on interaction movements driven by text prompts. 
Similar to navigation, we then augment the base model with a new object-aware transformer encoder and fine-tune this encoder separately. 

For the input to this branch at each denoising step $t$, we leverage Basis Point Sets (BPS) \cite{prokudin2019efficient} to calculate two key features: object geometry and the human-object relationship.
First, a sphere with a radius of $1.0$m is defined around the object's center, and $1024$ points are randomly sampled inside this sphere to form the BPS. The distance between each point in the BPS and the object's surface is then calculated, capturing the object's geometric features and stored as $\mathbf{B}_{O} \in \mathbb{R}^{1024}$.
Next, for each body pose $\mathbf{x}_t^n$ at timestep $n$ in the noisy input sequence, we calculate the minimum distance from each BPS point to any body joint, giving $\mathbf{B}^n \in \mathbb{R}^{1024}$. The resulting sequence of features $\mathbf{B}_H = \left[ \mathbf{B}^1, \dots, \mathbf{B}^N \right]$ represents the human-object relationship throughout the entire motion.
Finally, the object and human-object interaction features are concatenated with the original pose representation at each timestep $\left [ \mathbf{x}_t^n ;\; \mathbf{B}^n ;\; \mathbf{B}_O \right ]$ and fed to an MLP to generate a merged representation, which serves as the input to the scene-aware branch.

At test time, a collision objective is used to discourage penetrations between human and object. This is very similar to the collision loss described in \cref{t2loco}, but the SDF volume is computed for the 3D object and body vertices that are inside the object are penalized. Please see the supplementary material for details.

\begin{figure*}[t]
    \centerline{
    \includegraphics[width=\textwidth]{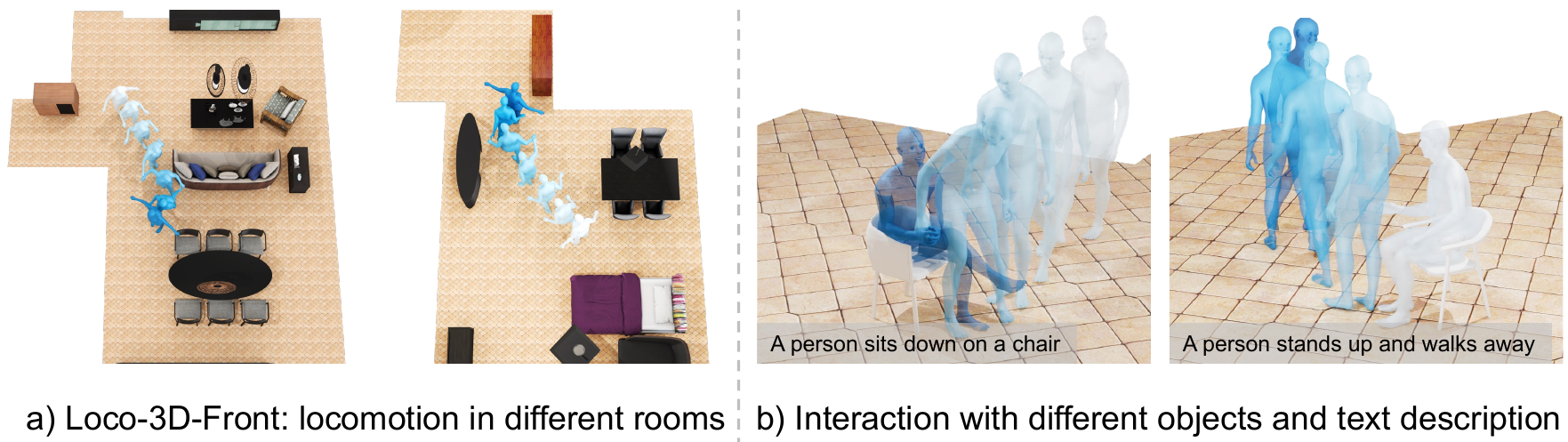}}
    \vspace{-2mm}
    \caption{ (\textbf{a}) Loco-3D-FRONT contains locomotion placed in 3D-FRONT \cite{fu20213d} scenes without collisions. (\textbf{b}) We augment SAMP \cite {21iccv_samp} by randomly selecting chairs from 3D-FRONT to match the motions and annotating a text description for each sub-sequence.}
    \vspace{-5mm}
    \label{fig:dataset}
\end{figure*}
\vspace{-2mm}
\paragraph{Scene-aware training and data.}
To train the scene-aware branch, we utilize the SAMP dataset \cite{21iccv_samp}, which captures motions and objects simultaneously. Specifically, we focus on ``sitting'' and ``stand-up'' interactions extracted from 80 sitting motion sequences in the SAMP dataset involving chairs of varying heights, as shown in \cref{fig:dataset}(b).
To diversify the object geometry, we randomly select objects from 3D-FRONT \cite{fu20213d} to match the contact vertices on human poses in the original SAMP motion sequences. This matching is achieved using the contact loss and collision loss techniques outlined in MOVER \cite{yi2022mover}. 

The original SAMP motions are often lengthy ($\sim$100 sec) and lack paired textual descriptions. For instance, a ``sit'' motion sequence involves walking to an object, sitting down, standing up, and moving away.
To effectively learn individual skills, we extract sub-sequences containing specific interactions that begin or end with a sitting pose, such as ``walk then sit'', ``stand up then sit'', ``stand up from sitting'', and ``walk from sitting.'' 
Furthermore, we annotate textual descriptions for each sub-sequence, which often incorporate the style of sitting poses, such as ``a person walks and sits down on a chair while crossing their arms.''
Applying left-right data augmentation to motion and objects results in approximately 200 sub-sequences for each motion sequence, each paired with corresponding text descriptions and featuring various objects.

%

\section{Experimental Evaluation}
\vspace{-2mm}
\subsection{Implementation Details}

\paragraph{Training.}
The scene-agnostic branch of our navigation model is trained on the 3D motions and text descriptions from the Loco-3D-FRONT dataset for 420k optimization steps. 
Subsequently, we freeze the base model weights and fine-tune the scene-aware branch, with additional 2D floor map inputs, for a further 20k steps. 
Similarly, the scene-agnostic base of our interaction model first trains on a mix of HumanML3D~\cite{22cvpr_humanml3d} and SAMP~\cite{21iccv_samp} data without objects for 400k steps. Then, the object-aware branch is fine-tuned on our text-annotated SAMP data with 3D object inputs for an additional 20k steps.
\vspace{-3mm}
\paragraph{Test-time guidance.}
For the navigation model, we set the guidance weight $\alpha$ to $30$ for goal-reaching guidance and $1000$ for collision guidance. In the interaction model, we utilize weights of $1000$ for goal-reaching loss and $10$ for the collision SDF loss. To ensure smooth generation results, we exclude the inference guidance at the final time step of denoising.
For a fair comparison with baselines, we do \emph{not} use inference guidance unless explicitly stated in the experiment.
\vspace{-3mm}
\subsection{Evaluation Data and Metrics}
\paragraph{Navigation.}
Navigation performance is assessed using the test set of Loco-3D-FRONT, comprising roughly 1000 sequences. 
Our metrics evaluate the generated root trajectory and the full-body motion after in-painting separately. 
For the root trajectory, we measure goal-reaching accuracy for the 2D (horizontal $xz$) root \textbf{position} (m), \textbf{orientation} (rad), and root \textbf{height} (m). 
The \textbf{collision ratio}, the fraction of frames within generated trajectories where a collision occurs, evaluates the consistency of root motions with the environment. 
For the full-body motion after in-painting, we use common metrics from prior work~\cite{22cvpr_humanml3d}. \textbf{FID} measures the realism of the motion, \textbf{R-precision} (top-3) evaluates consistency between the text and motion, and \textbf{diversity} is computed based on the average pairwise distance between sampled motions.
Additionally, the \textbf{foot skating} ratio~\cite{karunratanakul2023gmd} evaluates the physical plausibility of motion-ground interaction by the proportion of frames where either foot slides a distance greater than a specified threshold (2.5 cm) while in contact with the ground (foot height $<$5 cm).
\vspace{-3mm}
\paragraph{Interactions.}
To evaluate full-body human-object interactions, we use the established test split of the SAMP dataset~\cite{21iccv_samp}, which contains motions related to sitting.
Same as navigation, we analyze goal-reaching accuracy through position, orientation, and height errors. 
Furthermore, we assess physical plausibility by computing average \textbf{penetration values} and \textbf{penetration ratios} between the generated motion and interaction objects.
The penetration value is the mean SDF value across all interpenetrated body vertices of the generated motions, while the ratio is the fraction of generated poses containing penetrations (\ie, SDF values $<$ $-3$ cm) over all generated motion frames. 
We also perform a \textbf{user study} to compare methods. We employ Amazon Mechanical Turk (AMT)~\cite{mturk} to solicit assessments from 30 individuals. Raters are presented with two side-by-side videos of generated interactions and asked which is more realistic. Please see the supplementary material for more details.

\vspace{-2mm}
\subsection{Comparisons}
\vspace{-1mm}
\paragraph{Navigation.}
We conduct a comparative analysis of our method with previous scene-aware and scene-agnostic motion generation approaches, shown in \tref{table:trajectory_cmp}. 
Every method is conditioned on a text prompt along with a start and end goal pose, as described in \cref{t2loco}. The TRACE baseline and our method \modelname also receive the 2D floor map as input. 

\setlength{\tabcolsep}{6pt}
\begin{table*}[t!]
  \centering
  \caption{
Evaluation of navigation motion generation on the Loco-3D-FRONT test set. 
(\textbf{Left}) For generated pelvis trajectories, our approach achieves the best goal-reaching accuracy with low collision rate. 
(\textbf{Right}) After in-painting the full-body motion, our method maintains diverse and realistic motion that aligns with the given text prompt, competitive with diffusion-based scene-agnostic GMD and OmniControl.
}
  \label{table:trajectory_cmp}
  \vspace{-1mm}
  \resizebox{\columnwidth}{!}{
  \begin{tabular}{l cccc | cccc}
    \toprule
    \multicolumn{1}{l}{} & \multicolumn{4}{c}{\textbf{Root trajectory evaluation}} & \multicolumn{4}{c}{\textbf{Full-body motion evaluation}} \\
    \multicolumn{1}{l}{} & \multicolumn{3}{c}{Goal-reaching error $\downarrow$} & \multicolumn{1}{c}{} & & & & \\
    \multicolumn{1}{l}{\textbf{Method}} & \multicolumn{1}{c}{Pos.} & \multicolumn{1}{c}{Orient.} & \multicolumn{1}{c}{Height}  & \multicolumn{1}{c}{Collision $\downarrow$}  & FID $\downarrow$ & R-precision $\uparrow$ & Diversity $\uparrow$ & Foot skating $\downarrow$ \\ 
\midrule
Ground Truth & - & - & - & - & 0.010 & 0.672 & 7.553 & 0.000 \\
\midrule
GMD\cite{karunratanakul2023gmd}  &  0.374 & 1.231 & - & - & \textbf{13.160} & 0.114 & 4.488 & 0.181 \\
OmniContol\cite{xie2023omnicontrol} & 1.226 & 1.018 & 1.159 & - & 22.930 & \textbf{0.458}  & \textbf{7.128} & 0.094 \\
TRACE \cite{rempeluo2023tracepace} &  0.205  & 0.152  &0.010 & 0.055 & 22.669 & 0.144 & 6.501 & 0.058 \\
\midrule
Ours (1-stage train) & 0.197 & 0.132 & 0.013 & \textbf{0.028} & 22.372 & 0.152 & 6.347 & 0.062 \\

    Ours & \textbf{0.169} & \textbf{0.119} & \textbf{0.008} & 0.031 & 20.465 & 0.376 & 6.415 & \textbf{0.056} \\
\bottomrule
  \end{tabular}
  }
  \vspace{-2mm}
\end{table*}

\begin{figure*}[t!] 
    \includegraphics[width=\textwidth]{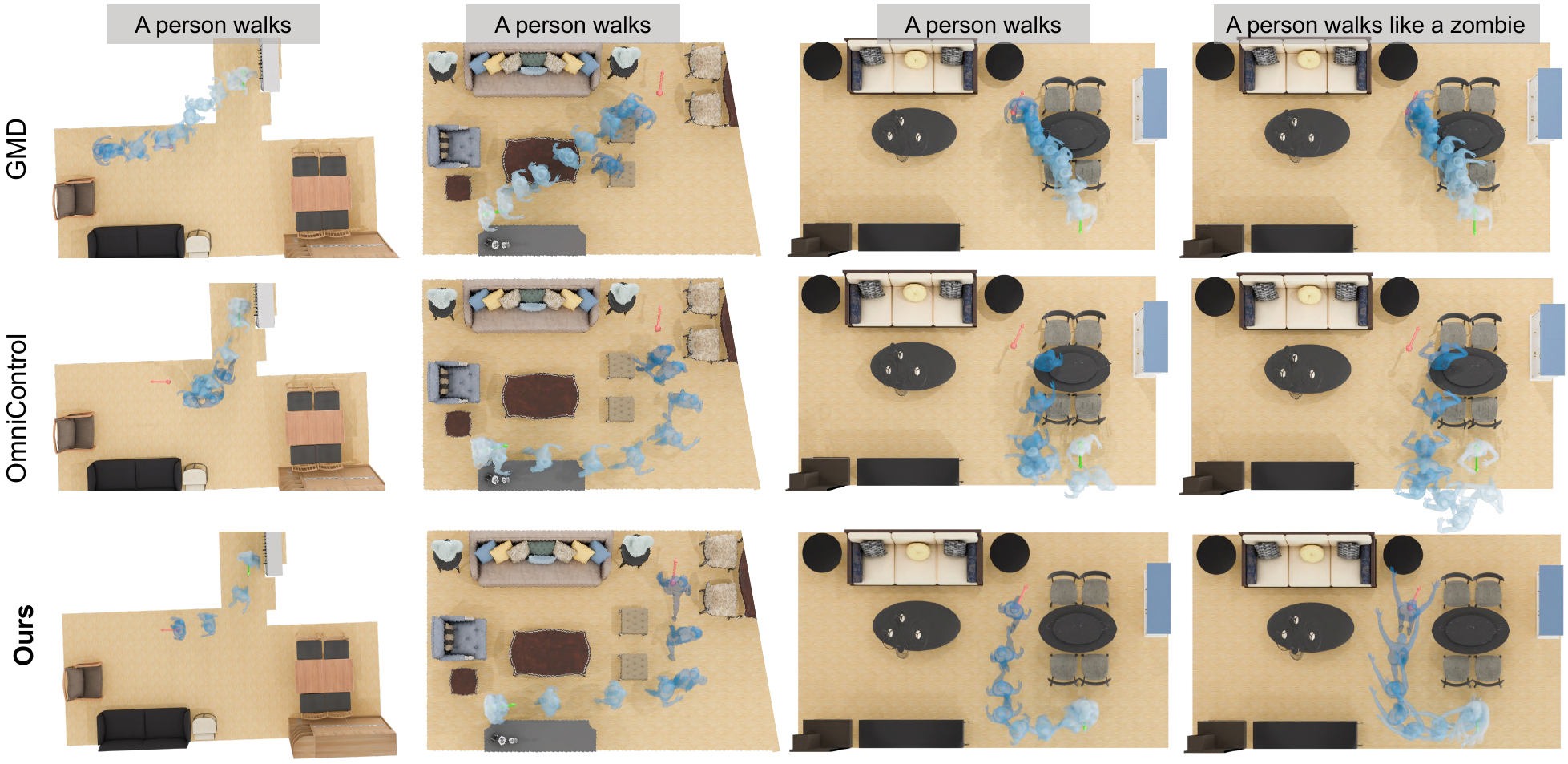}
    \vspace{-5mm}
    \caption{
    Navigation generation performance. 
    The start pose is the \textcolor{ForestGreen}{green} arrow, and the goal pose is the \textcolor{red}{red} arrow. Our method more accurately reaches the goal and avoids obstacles while style is controlled by a text prompt. 
    }
    \vspace{-5mm}
    \label{fig:cmp_loco_motion}
\end{figure*}

We first compare to GMD \cite{karunratanakul2023gmd} and OmniControl \cite{xie2023omnicontrol}, previous scene-agnostic text-to-motion diffusion models trained on HumanML3D to follow a diverse range of kinematic motion constraints.
GMD utilizes the horizontal pelvis positions $\left(x,z\right)$ of both the start and end goals to generate a dense root trajectory and subsequently the full-body motion. OmniControl takes as input the horizontal pelvis positions $\left(x,z\right)$ along with the height $y$ to directly generate full-body motion in a single stage.
Our navigation model achieves better goal-reaching accuracy, e.g., 16.9 cm for root position, since it is trained specifically for the goal-reaching locomotion task. 
More importantly, in the right half of \cref{table:trajectory_cmp} the full-body motion from our method after in-painting is comparable in terms of realism, text-following, and diversity, while achieving the best foot skating results. 
This demonstrates that our approach adds scene-awareness to locomotion generation, without compromising realism or text control.

To justify our two-branch model architecture, we adapt TRACE\cite{rempeluo2023tracepace}, a recent root trajectory generation model designed to take a 2D map of the environment as input. 
The adapted TRACE architecture is very similar to our model in \cref{fig:arctecture}(a), but instead of using a separate scene-aware branch, the base transformer directly takes the encoded 2D floor map features as input. 
This results in a single-branch architecture that must be trained from scratch, as opposed to our two-branch fine-tuning approach.
\cref{table:trajectory_cmp} reveals that our method generates more plausible root trajectories with fewer collisions and more accurate goal-reaching.
%
%
%
We also see that training our full two-branch architecture from scratch (\textit{1-stage train} in \cref{table:trajectory_cmp}), instead of using pre-training then fine-tuning, degrades both goal reaching and final full-body motion after in-painting. 

A qualitative comparison of generated motions in different rooms is shown in \fref{fig:cmp_loco_motion}. 
GMD tends to generate simple walking-straight trajectories. OmniControl and GMD do not reach the goal pose accurately and ignore the surroundings, leading to collisions with the environment.
%
Our method \modelname is able to generate diverse locomotion styles controlled by text in various scenes, achieving superior goal-reaching accuracy compared to other methods.

\setlength{\tabcolsep}{8pt}
\begin{table*}[t]
\centering
\caption{
Evaluation of human-object interaction motion generation on SAMP~\cite{21iccv_samp} sitting test set. Compared to DIMOS, our approach reaches the goal pose more accurately and exhibits fewer object penetrations, resulting in higher human preference. 
}
\label{table:interaction_cmp}
\vspace{-2mm}
\resizebox{\textwidth}{!}{
\begin{tabular}{l ccc cc | c}
\toprule
\multicolumn{1}{l}{} & \multicolumn{3}{c}{\textbf{Goal-reaching error} $\downarrow$} & \multicolumn{2}{c}{\textbf{Object penetration} $\downarrow$} & \textbf{User study}\\
\textbf{Method} & Pos. & Height &  Orient. & Value & \multicolumn{1}{c}{Ratio} & \textbf{preference} $\uparrow$ \\

\midrule
DIMOS \cite{zhao2023dimos} & 0.2020 & 0.1283 & 0.4731 & 0.0193 & 0.1076 & 29.1\% \\
Ours & \textbf{0.1445} &
\textbf{0.0120} &
\textbf{0.2410} & \textbf{0.0043} & \textbf{0.0611} & \textbf{71.9\%} \\ 

\bottomrule
\end{tabular}
}
\vspace{-5mm}
\end{table*}


\vspace{-2mm}
\paragraph{Interaction.}

\begin{figure*}[t!]
    \centerline{
    \includegraphics[width=\textwidth]{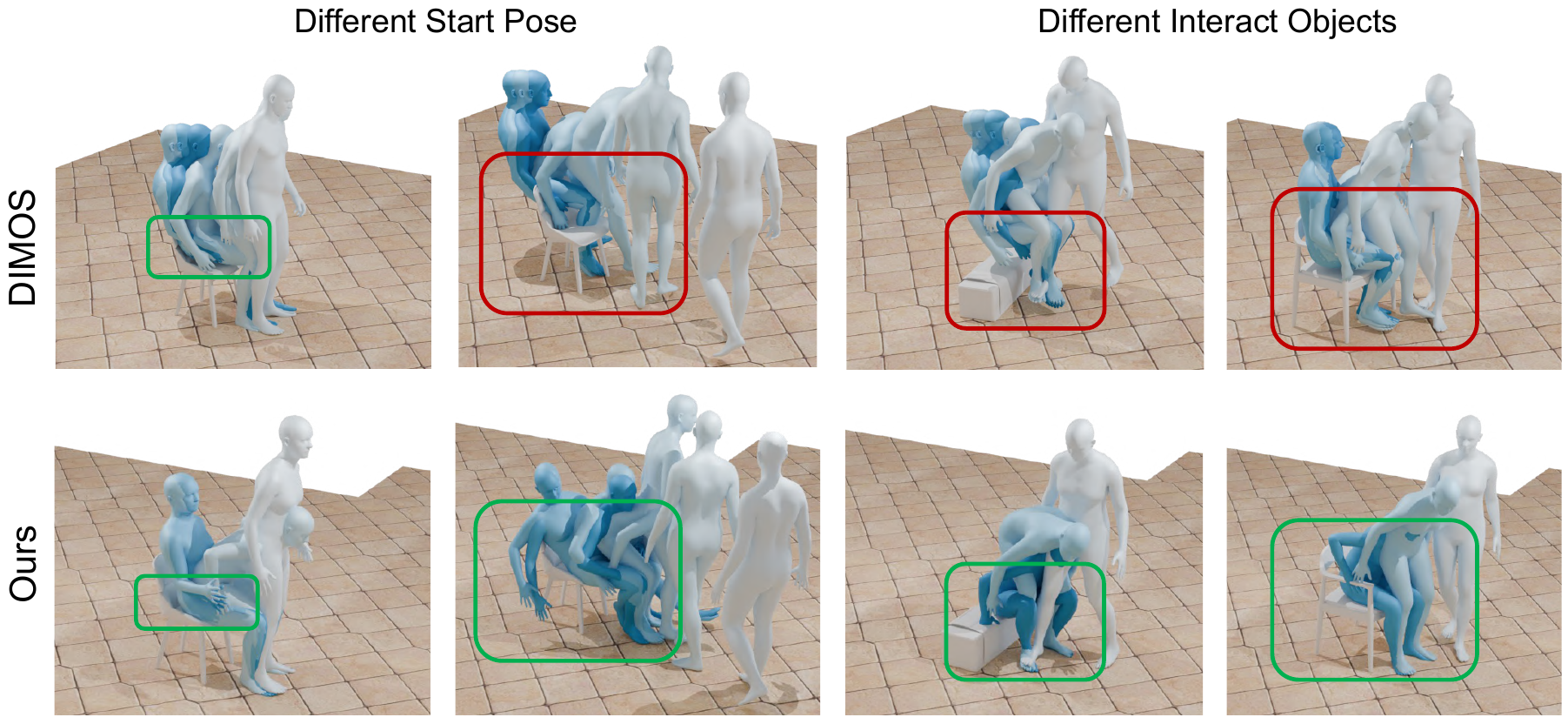}}
    \vspace{-2mm}
    \caption{Compared with DIMOS \cite{zhao2023dimos}, our method generates more realistic human-object interactions with reduced floating and interpenetrations.
    }
    \vspace{-5mm}
    \label{fig:interaction_comparison}
\end{figure*}

\tref{table:interaction_cmp} compares our approach to DIMOS~\cite{zhao2023dimos}, a state-of-the-art method to generate interactions trained with reinforcement learning. 
DIMOS requires a full-body final goal pose as input to the policy, unlike our approach which uses just the pelvis pose. 
Despite this, DIMOS struggles to reach the goal accurately, likely due to error accumulation during autoregressive rollout.
Our method showcases fewer instances of interpenetration with interaction objects and the user study reveals a distinct preference for motions generated by our approach (preferred 71.9\%) over those produced by DIMOS.
%
%
\fref{fig:interaction_comparison} compares the approaches qualitatively, where we see that more accurate goal-reaching reduces floating or penetrating the chair during sitting. Moreover, the interactions generated by DIMOS lack diversity, and cannot be conditioned on text.
\setlength{\tabcolsep}{6pt}
\begin{table*}[t]
\centering
\caption{
Test-time guidance evaluation. Adding guidance to reach goal poses and avoid collisions during inference improves performance. Lower is better for all metrics. 
}
\label{table:guidance_ablation}
\vspace{-2mm}
\resizebox{\textwidth}{!}{
\begin{tabular}{cc | cc | ccc}

\toprule
\multicolumn{2}{c}{\textbf{Guidance}} & \multicolumn{2}{c}{\textbf{Navigation}} & \multicolumn{3}{c}{\textbf{Interaction}} \\
Goal Reach & \multicolumn{1}{c}{Collision} & Goal Pos. & \multicolumn{1}{c}{Collision} & Goal Pos. & Pen. Val. & Pen. Ratio \\
\midrule
{\color{red}\xmark} & {\color{red}\xmark} & 0.1568 & 0.0294 & 0.1445  & 0.0043 & 0.0611 \\ 
{\color{ForestGreen}\checkmark} & {\color{red}\xmark} & \textbf{0.118} &  0.0342 & 0.1453 & 0.0050 & 0.0554\\
{\color{red}\xmark} & {\color{ForestGreen}\checkmark} &  0.1550 & 0.0013 & 0.1407 & \textbf{0.0040} & \textbf{0.0414} \\
{\color{ForestGreen}\checkmark} & {\color{ForestGreen}\checkmark} & 0.1241 & \textbf{0.0012} & \textbf{0.1404} & 0.0045 & 0.0494 \\
\bottomrule
\end{tabular}
}
\vspace{-3mm}
\end{table*}


\vspace{-3mm}
\subsection{Analysis of Capabilities}
\begin{figure}[ht!]
    \centerline{
    \includegraphics[width=\columnwidth]{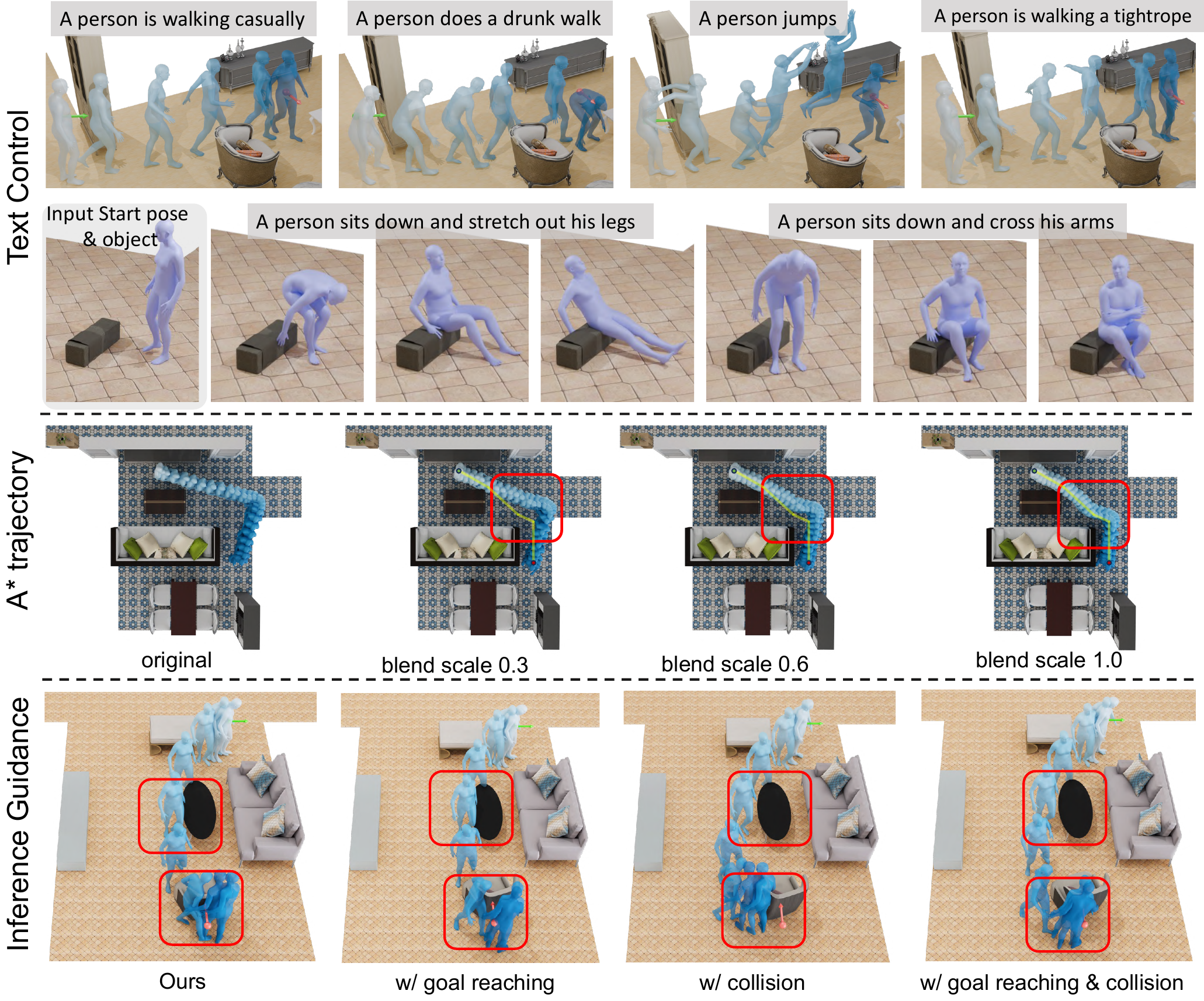}}
    \vspace{-2mm}
    \caption{ \modelname capabilities. (\textbf{Top}) Diverse text control; (\textbf{Middle}) Following A* path with adherence controlled by the blend scale; 
(\textbf{Bottom}) Test-time guidance encourages locomotion to reach the goal accurately without colliding with the environment.
}
    \vspace{-5mm}
    \label{fig:capabilities}
\end{figure}

In \fref{fig:teaser}, our method carries out a sequence of actions, enabling traversal and interaction with multiple objects within a scene. 
\cref{fig:capabilities} demonstrates additional key capabilities. 
In the top section, our method is controlled through a variety of text prompts. 
For interactions in particular, diverse text descriptions disambiguate between actions like sitting or standing up, and allow stylizing the sitting motion, e.g., with crossed arms. 
In the middle section, we enable user control over trajectories by adhering to a predefined A* path. By adjusting the blend scale, users can adjust how closely the generated trajectory follows A*.
At the bottom of \cref{fig:capabilities}, we harness guidance at test time to encourage motions to reach the goal while avoiding collisions and penetrations. As shown in \cref{table:guidance_ablation}, combining guidance losses gives improved results both for navigation and interactions. 

\vspace{-3mm}
\section{Discussion}
We introduced \modelname, a novel method for text-controlled scene-aware motion generation. 
By first pre-training a scene-agnostic text-to-motion diffusion model on large-scale motion capture data and subsequently fine-tuning with a scene-aware component, our text-conditioned method enables generating realistic and diverse human-object interactions within 3D scenes. 
To support such training, we introduced the new Loco-3D-FRONT dataset containing realistic navigation motions placed in 3D scenes, and extended the SAMP dataset with additional objects and text annotations.
Experiments demonstrate that our generated motion is on par with state-of-the-art diffusion models, while improving the plausibility and realism of interactions compared to prior work.
\vspace{-3mm}
\paragraph{Limitations \& Future Work.}
While our navigation model enables accurate goal-reaching and text-to-motion controllability, the two-stage process can sometimes lead to a disconnect between the generated pelvis trajectory and in-painted full-body poses. Exploring new one-stage models, capable of simultaneously generating pelvis trajectories and poses, would streamline the process. Additionally, our current approach, which operates on 2D floor maps, restricts the ability to handle intricate interactions, such as a person stepping over a small stool.

Our current approach is aimed at controllability to allow users to specify text prompts or goal objects and locations. However, our method may also fit into recently proposed pipelines~\cite{xiao2023unified} that employ LLM planners to specify a sequence of actions and contact information that could be used to guide our motion generation. 
Looking ahead, we also aim to broaden the spectrum of actions modeled by the system, to encompass activities such as lying down and touching. 
Furthermore, enabling interactions with dynamic objects will allow for more interactive and realistic scenarios.



\mheading{Acknowledgments}
{\small
Thanks Yangyi Huang and Yifei Liu for technical support.
Thanks Mathis Petrovich and Nikos Athanasiou for the fruitful discussion about text-to-motion synthesis.
Thanks Tomasz Niewiadomski, Taylor McConnell, and Tsvetelina Alexiadis for running the user study.
}

{\small \qheading{Disclosure}
MJB has received research gift funds from Adobe, Intel, Nvidia, Meta, and Amazon. MJB has financial interests in Amazon, Datagen Technologies, and Meshcapade GmbH.  While MJB is a consultant for Meshcapade, his research in this project was performed solely at, and funded solely by, the Max Planck Society.
}


\begin{appendix}
\section{Ablation Study}
\setlength{\tabcolsep}{6pt}
\begin{table*}[hp]
  \centering
  \caption{
Ablation study comparing various full-body infilling methods and different representations of navigation motion generation using the Loco-3D-FRONT test set.
(\textbf{Left}) For generated pelvis trajectories, our approach achieves the best goal-reaching accuracy with low collision rate. 
(\textbf{Right}) After in-painting the full-body motion, our method preserves diverse and realistic movements that align with the provided text prompt, much like the model employing an alternative OminiControl full-body inpainting technique. However, our approach distinctly outperforms the model utilizing full-body representation.
}
  \label{table:sup_trajectory_cmp}
  \vspace{-1mm}
  \resizebox{\columnwidth}{!}{
  \begin{tabular}{l cccc | cccc}
    \toprule
    \multicolumn{1}{l}{} & \multicolumn{4}{c}{\textbf{Root trajectory evaluation}} & \multicolumn{4}{c}{\textbf{Full-body motion evaluation}} \\
    \multicolumn{1}{l}{} & \multicolumn{3}{c}{Goal-reaching error $\downarrow$} & \multicolumn{1}{c}{} & & & & \\
    \multicolumn{1}{l}{\textbf{Method}} & \multicolumn{1}{c}{Pos.} & \multicolumn{1}{c}{Orient.} & \multicolumn{1}{c}{Height}  & \multicolumn{1}{c}{Collision $\downarrow$}  & FID $\downarrow$ & R-precision $\uparrow$ & Diversity $\uparrow$ & Foot skating $\downarrow$ \\ 
\midrule

Ours (OmniControl \cite{xie2023omnicontrol} in-painting) &  0.459 & 0.999 & 0.090 & 0.073 & \textbf{17.927} & \textbf{0.396} & 6.288 & \textbf{0.0308} \\

Ours (full-body rep) & 0.844 & 0.016 & 0.110 & 0.124 & 24.642 & 0.189 & \textbf{6.967} & 0.169 \\

    Ours & \textbf{0.169} & \textbf{0.119} & \textbf{0.008} & \textbf{0.031} & 20.465 & 0.376 & 6.415 & 0.056 \\
\bottomrule
  \end{tabular}
  }
\end{table*}

\paragraph{Alternative Full-Body In-painting Approach.}

While our root trajectory generation approach can integrate with several motion in-painting techniques, in the main paper we use PriorMDM~\cite{priormdm}. As an alternative, we evaluate our method using OminiControl \cite{xie2023omnicontrol} for in-painting in \cref{table:sup_trajectory_cmp}. However, OmniControl overrides our generated dense pelvis trajectory and jointly generates full-body locomotion with a new pelvis trajectory. This severely degrades the goal-reaching ability (from $0.169$ cm to $0.459$ cm) as demonstrated in Table \ref{table:sup_trajectory_cmp}. Therefore, we choose to utilize PriorMDM as our body motion in-painting method. It aligns well with our generated trajectory, resulting in the generation of plausible locomotion while maintaining adherence to the goal position.

\paragraph{One-stage Navigation Motion Generation.}

To evaluate the efficacy of our two-stage navigation model design, we compare to a single-stage full-body motion generation ablation of our model. This model operates on the same input data but directly generates full-body locomotion. However, as shown in \cref{table:sup_trajectory_cmp}, this approach limits goal-reaching ability and does not produce motion styles that align with the input text. 
The local poses are somewhat dissociated from the global pelvis trajectories, allowing for trajectory variations while maintaining the same motion style.
For instance, individuals can walk along different paths while maintaining consistency in their motion style.

\section{Details on User Study for Interaction Motions}
\begin{figure*}[t]
    \centerline{
    \includegraphics[width=\textwidth]{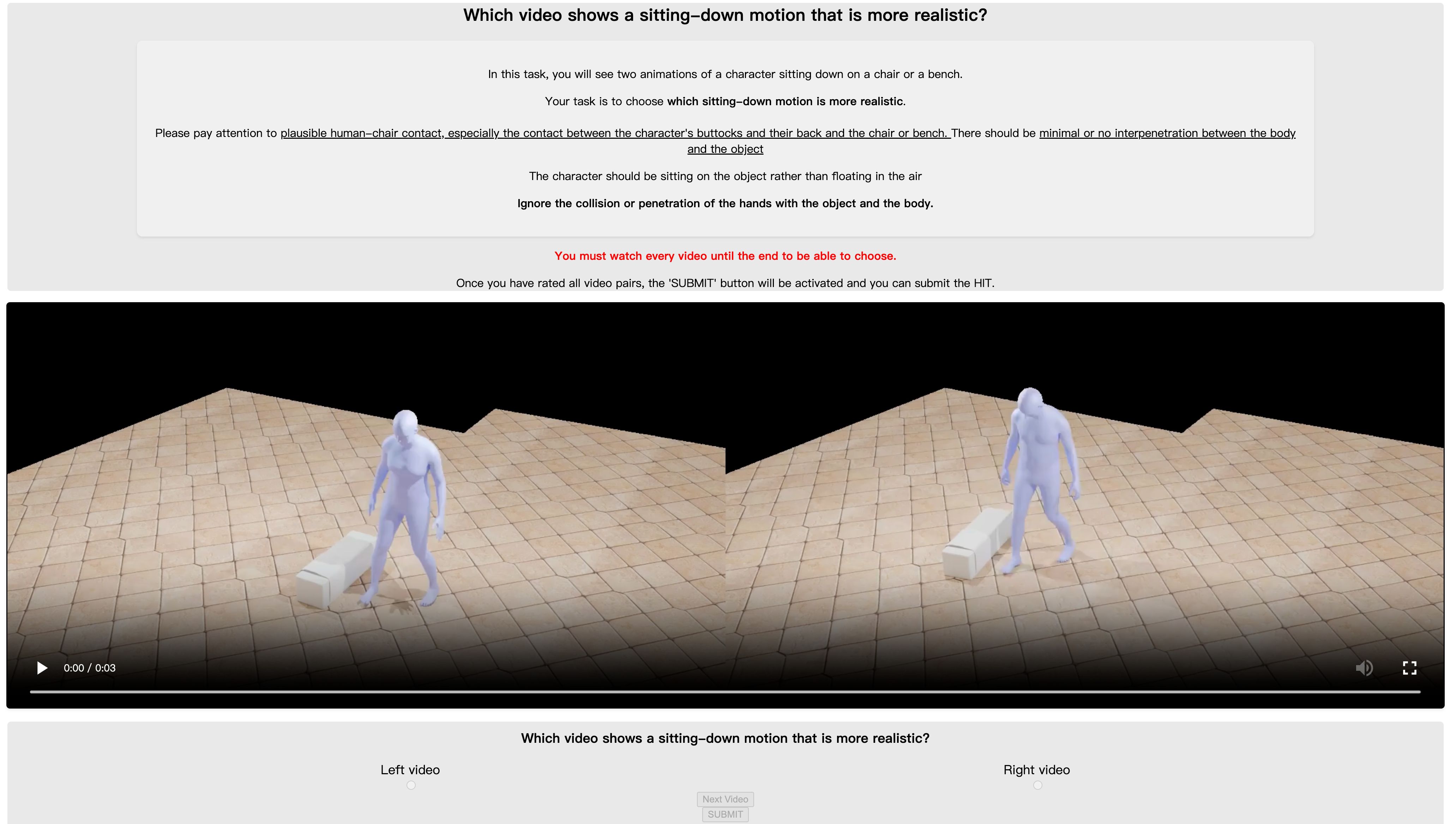}}
    \caption{The layout of our perceptual study for evaluating the plausibility of human-object interaction.}
    \label{fig:user_study}
\end{figure*}
To evaluate the plausibility of human-object interaction, we perform a user study to compare our method and DIMOS \cite{zhao2023dimos}. 
We employ Amazon Mechanical Turk (AMT) \cite{mturk} to solicit assessments from 30 individuals. 
Raters are presented with two side-by-side videos depicting generated interactions and asked to determine which appeared more realistic, particularly focusing on the contact between the character's buttocks and their back with the chair or bench, and the presence of minimal or no interpenetration between the body and the object. 
We present 70 test videos with the positions of our generated videos and DIMO's results randomly shuffled horizontally.
In order to filter out poor responses, we duplicated our 5 test examples where clear preferences between two video results were evident, serving as catch trials. Ultimately, we obtained 65 useful responses out of 70 raters.
The full survey page is illustrated in \fref{fig:user_study}.
The user study reveals a distinct preference for motions generated by our approach (preferred 71.9\%) over those produced by DIMOS.

\section{Details on Collision Guidance Used in Interaction Motion Generation}
At test time, a collision objective is used to discourage penetrations between humans and objects.
Remarkably, our interaction motion generation model outputs 3D joint positions. We then link randomly sampled vertices on the SMPL mesh surfaces with the 3D skeletons in an A-pose, allowing us to obtain the posed sampled vertices for each new pose.
This is defined as $\mathcal{J}_c = \text{SDF}(\hat{\mathbf{x}}_0, \mathcal{S_{O}})$ where $\text{SDF}$ calculates the SDF volume of the object $O$, then queries the sign distance value at each time step of the body vertices. Positive distances, indicating body vertices inside the interactive object, are averaged to get the final loss.

\end{appendix}



%
%
\bibliographystyle{configs/splncs04}
\bibliography{ref}
\end{document}